\documentclass[conference]{IEEEtran}
\IEEEoverridecommandlockouts
\usepackage{cite}
\usepackage{amsmath,amssymb,amsfonts}
\usepackage{algorithmic}
\usepackage{graphicx}
\usepackage{textcomp}
\usepackage{subfigure}
\usepackage{caption}

\usepackage{multirow,balance}

\usepackage{xcolor}
\def\BibTeX{{\rm B\kern-.05em{\sc i\kern-.025em b}\kern-.08em
    T\kern-.1667em\lower.7ex\hbox{E}\kern-.125emX}}
\begin{document}

\title{Deep Cervix Model Development from Heterogeneous and Partially Labeled Image Datasets\\
}

\author{\IEEEauthorblockN{Anabik Pal}
\IEEEauthorblockA{\textit{National Library of Medicine} \\
\textit{National Institutes of Health (NIH)}\\
Bethesda, Maryland, USA \\
anabikpal@gmail.com}
\and
\IEEEauthorblockN{Zhiyun Xue}
\IEEEauthorblockA{\textit{National Library of Medicine} \\
\textit{National Institutes of Health (NIH)}\\
Bethesda, Maryland, USA \\
zhiyun.xue@nih.gov}
\and
\IEEEauthorblockN{Sameer Antani}
\IEEEauthorblockA{\textit{National Library of Medicine} \\
\textit{National Institutes of Health (NIH)}\\
Bethesda, Maryland, USA \\
sameer.antani@nih.gov}
}

\maketitle

\begin{abstract}
Cervical cancer is the fourth most common cancer in women worldwide. The availability of a robust automated cervical image classification system can augment the clinical care provider’s limitation in traditional visual inspection with acetic acid (VIA). However, there are a wide variety of cervical inspection objectives which impact the labeling criteria for criteria-specific prediction model development. Moreover, due to the lack of confirmatory test results and inter-rater labeling variation, many images are left unlabeled.

Motivated by these challenges, we propose a self-supervised learning (SSL) based approach to produce a pre-trained cervix model from unlabeled cervical images. The developed model is further fine-tuned to produce criteria-specific classification models with the available labeled images. We demonstrate the effectiveness of the proposed approach using two cervical image datasets. Both datasets are partially labeled and labeling criteria are different. The experimental results show that the SSL-based initialization improves classification performance (Accuracy: 2.5\%↑)  and the inclusion of images from both datasets during SSL further improves the performance (Accuracy: 1.5\%↑). Further, considering data-sharing restrictions, we experimented with the effectiveness of Federated SSL and find that it can improve performance over the SSL model developed with just its images. This justifies the importance of SSL-based cervix model development. We believe that the present research shows a novel direction in developing criteria-specific custom deep models for cervical image classification by combining images from different sources unlabeled and/or labeled with varying criteria, and addressing image access restrictions.

\end{abstract}


\section{Introduction}\label{sect::intro}
Cervical cancer is the fourth common cancer in women worldwide. Regular cervical screening can help in the early detection of the pre-cancerous lesions and reduce premature death. Visual inspection with acetic acid (VIA) is a commonly used low-cost cervical screening approach, but it suffers from significant inter- and intra-reader variability. In this regard, the availability of an automated cervical image classification system could address this limitation ~\cite{Liming2019}. It is challenging to develop a diagnostic system using traditional classification approaches as it needs enormous effort to develop effective hand-crafted features and are known to underperform~\cite{Srinivasan2006,Kim2013,Fernandes2018}. Hence, the deep learning approaches~\cite{Litjens2017} can be considered to build a robust classification model. However, the development of a deep model for robust classification needs a huge number of images and their class labels~\cite{Ching2018}. Labeling cervical images is costly, needs multiple experts' agreement and requires multiple diagnostic information. Transfer learning~\cite{yang2020}, i.e. transferring knowledge from natural images, is a commonly used approach to overcome these data limitations. However, transferring knowledge from the same domain may be more effective than transferring knowledge from a different domain.


Collaboration among multiple organizations to perform centralized learning (CL) by uniting all labeled images from all sources is another effective solution to overcome  the data scarcity. However, supervised learning using combined images has limitations. First, different imaging devices may be used for image acquisition resulting in variations in the visual quality across different sources. It is challenging to estimate the proper ratio of images to be selected from various sources for the CL. Moreover, there may be data-sharing restrictions. To deal with data-sharing restrictions, federated learning (FL) can be employed~\cite{Yang2019}. However, federated supervised learning needs research efforts to address variability in class distribution among the sources~\cite{Yang2020Fed}. Furthermore, inclusion of a dataset having noisy labels impede the training to produce robust models. Finally, and most importantly, the cervical image labeling criteria vary and depends on: availability of other diagnostic results, population under study, treatment planning, severity grading strategy, etc. The variety in the image labeling criteria across datasets makes the task more challenging as it restricts researchers in performing any kind of supervised collaborative (CL or FL)  learning.

In this paper, we propose a self-supervised learning (SSL) based approach to develop a pre-trained cervix model (or cervix model). We use two cervical image datasets which are (i) labeled in a heterogeneous manner: labeling criteria vary across datasets and (ii) partially labeled: not all images in the datasets are labeled. As the SSL does not require any label, it allows us to include all available images in our datasets for cervix model development. Both centralized SSL and federated SSL are experimented. To evaluate the effectiveness of the developed cervix model, criteria (or dataset) specific classification models are trained with the available labeled images. The classification networks are initialized with the developed cervix models. Note that, according to our survey, no image dataset is publicly available for machine learning research towards supporting experts' effort in visual assessment of acetic acid applied cervix and no well-accepted classification network is available for the present task. Hence, we confined our experiments only with the present datasets for the chosen competing algorithms.

In summary, our work is motivated from~\cite{Zhou2021} and~\cite{Chen2020}. In~\cite{Zhou2021}, for medical image representation, an encoder-decoder based architecture was used to reconstruct original images from synthetically distorted images. In~\cite{Chen2020}, a contrastive learning-based framework is proposed for natural image representation. In contrast to these works, we utilize the framework presented in ~\cite{Chen2020} and demonstrate the power of this framework for cervical image representation tasks in both centralized and federated learning schemes. We believe that our work has two key novelties: (a) it is the first work attempting to develop a cervix model from unlabelled images; (b) the first work where Federated Self-Supervised Learning  (FSSL) is demonstrated for any medical image representation.

The remainder of the paper is organized as follows: Section~\ref{sect::Method} discusses the proposed methodology. The experimental protocol and analysis of experimental results are presented in Section~\ref{sect::Experimental Protocol} and Section~\ref{sect::Experiments} respectively. Finally, Section~\ref{sec: Conclusion} concludes the paper.

\section{Methods}\label{sect::Method}

\subsection{Self-supervised learning}\label{sect::SSL}

Self-supervised learning (SSL) is a discriminative approach for visual representation learning. Developers define a pretext task and develop a deep model which captures the image semantics (i.e. good initialization weights for related domain's downstream tasks) with zero labeling cost. In this paper, we employ a contrastive feature learning algorithm as the pretext task- i.e. all images will be semantically well separated, and an image and its augmented version will be semantically closer~\cite{Chen2020}. During SSL training, every mini-batch of size $2N$ is constructed with $N$ random images and an augmented version of them. The training loss ($L_{i,j}$) between an image $i$ and its augmented version $j$ is given as:
\begin{equation}
    L_{i,j} = -log\frac{exp(f_i^Tf_j/ \tau)}{\sum_{i=1}^{2N}1_{[k\neq i]}exp(f_i^Tf_k/\tau)}
\end{equation}

where $f_i$ is the feature vector of $i^{th}$ image; $1_{[k\neq i]} \in \{0, 1\}$ is an indicator function evaluating to 1 iff $k\neq i$; $^T$ representation transpose operation and $\tau$ is a constant. The loss in a mini-batch is computed across all pairs constructed with an image and its augmented version.

\subsection{Federated Self-supervised Learning (FSSL)}\label{sect::FL}
Recently, several different data protection laws and regulations (GDPR 2018 by EU, CCPA 2020 in the US, etc) have been created to protect information leakage due to sensitive data sharing (especially medical/banking domain). This motivates the researchers to develop federated learning (FL) algorithms for robust inter-institutional collaborated deep development from the data distributed in multiple institutions (clients) without sharing the raw data~\cite{Yang2019}. 

In this paper, we experiment with Federated Self-supervised Learning (FSSL) for cervix model development. We perform FSSL in the following two different ways:- (a) Client-Server FSSL (CSFSSL) and (b) Peer-to-Peer FSSL (PPFSSL). The CSFSSL contains multiple clients and a server. In this framework, firstly, the server sends a deep model to all clients. Then independently, at every client, the model is fine-tuned for $E$-epochs with local data. After that, the updated models from all clients are aggregated at the server and used as the initialized model for the next iteration. This procedure continues until the deep model is trained. Note that, we use weight averaging for model aggregation. On the other hand, in the PPFSSL, SSL is circularly performed- firstly the starting client runs SSL with its own data for $E$-epochs and then sends the trained model to the next client at which SSL is performed for $E$-epochs using the received model as starting point and then the updated model is sent to next client and so on. This circular computation is performed for few iterations and the final model is shared among the clients.

\subsection{Proposed Approach: Cervical Model Development}\label{sect::Framework}
\begin{figure}[htp]
\centering
 {\includegraphics[width=.49\textwidth]{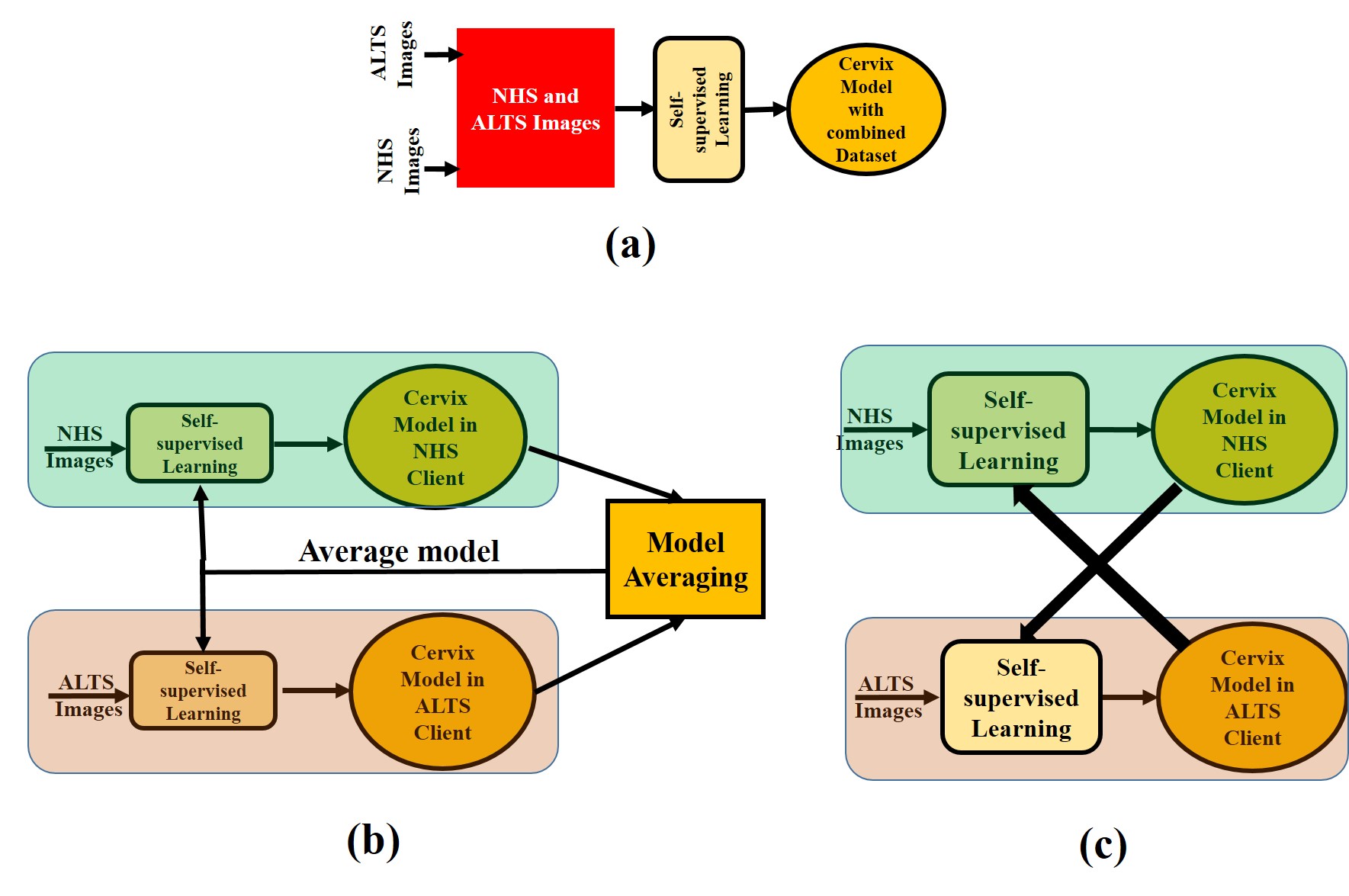}} \\
 \caption{Block diagram of the SSL training system-  (a) Centralized SSL (CSSL) (b) Client-Server FSSL (CSFSSL) and (c) Peer-to-Peer FSSL (PPFSSL).}
\label{fig:SSLBlockDiagram}
\end{figure}

In this paper, we use centralized SSL (CSSL) and two different variants of FSSL presented in Sec~\ref{sect::FL} for cervix model development. The block diagram of the CSSL, CSFSSL, and PPFSSL are shown in Fig~\ref{fig:SSLBlockDiagram} (a), Fig~\ref{fig:SSLBlockDiagram} (b) and Fig~\ref{fig:SSLBlockDiagram} (c) respectively. All methods use contrastive feature learning-based SSL presented in Sec~\ref{sect::SSL} for cervical model development. We perform flip (horizontal and vertical) axis, rotation, random shift, and random zoom, gamma changing, and brightness changing to get the augmented version of an image.

\section{Experimental Protocol}\label{sect::Experimental Protocol}
\subsection{Dataset description}\label{sect::Dataset}
National Cancer Institute (NCI) at the US National Institutes of Health (NIH) conducted two different cohort studies (NHS~\cite{Bratti2004} and ALTS~\cite{Schiffman2000}) for cervical examination. During these studies, in every visit, two images of the acetic acid applied cervix are captured. For the present research, NCI provides us a subset of images collected during these studies. The dataset containing NHS images is referred to as NHS dataset and the dataset containing ALTS images is referred to as ALTS dataset. NCI scientists labeled a subset of images available in NHS and ALTS datasets as case (disease) or control (non-disease) based on the availability of several screening and diagnostic information (like visual assessment, HPV, cytology, histopathology, colposcopy, etc). Different criteria were used to label these two datasets as the study objectives differs\footnote{NHS was from general population study, and ALTS was from triage study in colposcopic clinics.}. For our research, we split both NHS and ALTS datasets at the woman level into three disjoint subsets- train, validation, and test. The training images are used for model training, the validation images are used for training hyper-parameter selection, and the test images are used for classification performance evaluation. The split-wise number of patients, number of labeled images, and total available images (including labeled images) for the datasets are given in Table~\ref{table:dataset}.

\begin{table}[]
\caption{Data set splits.}
\center
\begin{tabular}{|c|c|c|c|c|c|c|c|}
\hline
\multirow{2}{*}{Split} & \multirow{2}{*}{Class} & \multicolumn{2}{c|}{Patients} & \multicolumn{2}{c|}{Labeled   Images} & \multicolumn{2}{c|}{Total Images}             \\ \cline{3-8} 
&    & NHS &ALTS                & NHS &ALTS  & NHS                   & ALTS                  \\ \hline
\multirow{2}{*}{Train} & Case                   & 91                 & 124                & 182                     & 248                    & \multirow{2}{*}{2029} & \multirow{2}{*}{3145} \\ \cline{2-6}
                       & Control                & 181                & 242                & 361                     & 481                    &                       &                       \\ \hline
\multirow{2}{*}{Valid} & Case                   & 22                 & 31                 & 44                      & 62                     & \multirow{2}{*}{520}  & \multirow{2}{*}{791}  \\ \cline{2-6}
                       & Control                & 45                 & 60                 & 90                      & 120                    &                       &                       \\ \hline
\multirow{2}{*}{Test}  & Case                   & 25                 & 34                 & 49                      & 68                     & \multicolumn{2}{c|}{}      \\ \cline{2-6}& Control   & 50                 & 65                 & 99                      & 130                    & \multicolumn{2}{c|}{}                                             \\ \hline
\end{tabular}
\label{table:dataset}
\end{table}

\subsection{Network architecture}\label{Sect::Network}
We use ResNet-50 as a backbone network architecture. For SSL approaches, firstly, the top 1000-way classification layer is removed and a dense layer with a ReLU activation is added. The dense layer serves as the image representation vector. We vary the number of neurons among [64, 128, 256] for this layer and empirically decide to set it $64$ as we obtain very close performances. For preparing the classification model, we remove the dense layer and put a single output neuron with sigmoid activation. The output obtained from the sigmoid layer is the case probability. The classification network is fully fine-tuned with the prepared cervix model.

\subsection{Competing methods}\label{Sect::Competing methods}
The paper aims to develop a cervix model, i.e., weight initialization method for cervical image classification network. We compare among six different approaches (i) \textbf{Random}: network weights are randomly initialized. (ii) \textbf{ImageNet}: network weights are taken from the pre-trained ImageNet classification model- knowledge is transferred from natural images. (iii) \textbf{Self-supervised Learning (SSL)}: SSL is employed with the available images in a dataset. (iv) \textbf{Centralized Self-supervised Learning (CSSL)}: All images from both datasets are combined to train the SSL as shown in Fig~\ref{fig:SSLBlockDiagram}(a). (v) \textbf{Client-Server Federated Self-supervised Learning (CSFSSL)}: Images are not shared, client SSL models are aggregated in the server (see Fig~\ref{fig:SSLBlockDiagram}(b)). (vi) \textbf{Peer-to-Peer Federated Self-supervised Learning (PPFSSL)}: Image are not shared, circularly SSL based fine-tuning is performed among the clients (see Fig~\ref{fig:SSLBlockDiagram}(c)).

\subsection{Parameter Settings}\label{Sect::Parameter}
We vary the network hyper-parameters for both SSL and classification model training and choose the best hyper-parameters based on the validation loss. The SSL networks are trained with following parameters: learning rate $0.01$, weight decay = $1e-5$, momentum = $0.9$, $\tau = 0.1$ (See Eq 1), epochs = $50$ and the classification networks are trained with following hyper-parameters: learning rate $0.001$, weight decay = $1e-6$, momentum = $0.09$, epochs = $50$, batch size =4. In both SSL and classification model training, we randomly shuffle images during batch construction. The learning stops when the validation loss is not decreasing for $5$ epochs. We use reverse class weighting to address the class imbalance issue in classification model training.

\subsection{Implementation}\label{Sect::Implementation methods}
The Keras~\cite{Chollet2015} deep learning tool-kit is used for implementing the networks. The networks are trained with 2 GeForce RTX 2080 Ti GPUs installed with an Intel(R) Xeon(R) Gold 5218 CPU (@ 2.30GHz). We implement federated learning in the same computing resources as sequential processes.

\subsection{Evaluation metrics}\label{sect::Evaluation metric}
In this paper, we evaluate the performance of the dataset-specific classification models produced from the supervised learning with the varying initialization approaches mentioned in Sec~\ref{Sect::Competing methods}. The following four quantitative evaluation metrics are computed for performance comparison: (1) Accuracy (ACC), (2) Recall, (3) Precision, (4) F1-Score. 

\section{Experimental results and discussion}\label{sect::Experiments}

\begin{figure}[htp]
\centering
 {\includegraphics[width=.22\textwidth]{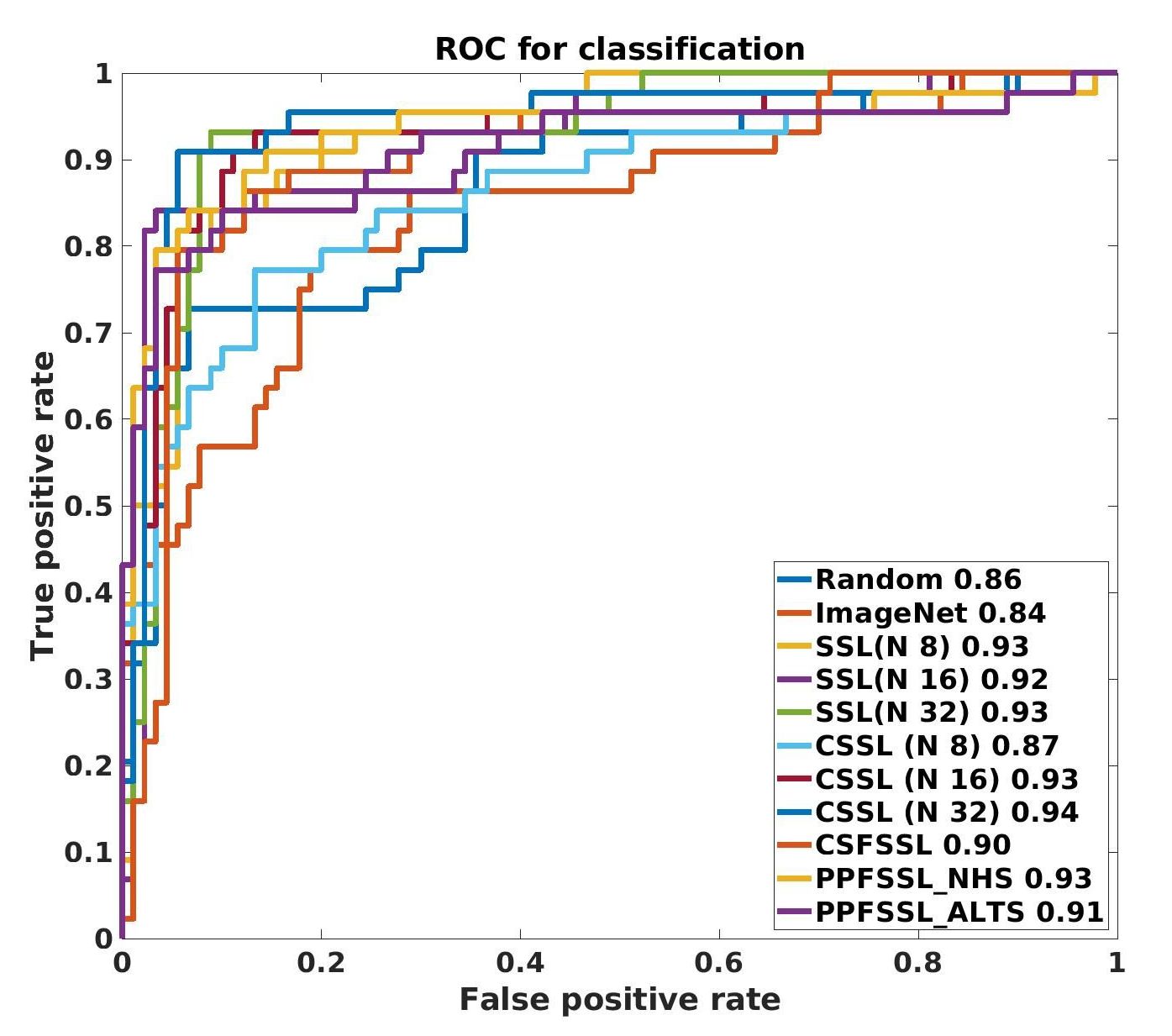}}  {\includegraphics[width=.22\textwidth]{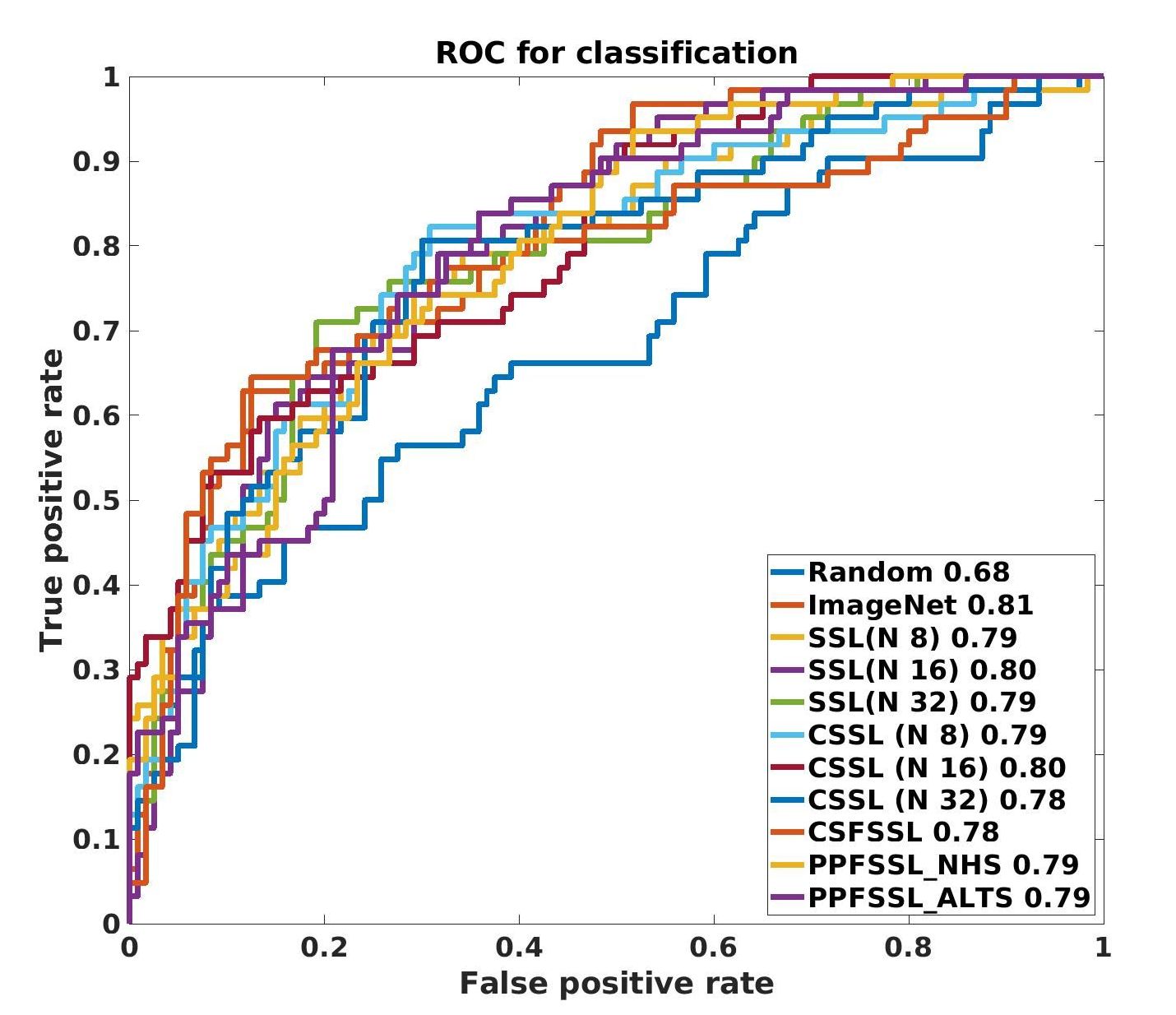}}\\
(a)\hspace{3.1cm}(b)
 \caption{Receiver Operating Curve (ROC): (a) NHS and (b) ALTS. The numeric values represent the AUC values for the classifiers with considered initialization.}
\label{fig:ROC}
\end{figure}

The Receiver Operating Curves (ROC) for all supervised classifiers for both NHS and ALTS datasets developed with varying initialization approaches are shown in Fig~\ref{fig:ROC}. The quantitative classification performance (performed with the metrics mentioned in Sec~\ref{Sect::Competing methods}) for the same is shown in Table~\ref{Table:Performance}. According to Table~\ref{Table:Performance}, we find that ImageNet initialization improves the accuracy for both datasets but recalls did not improve. The cervical image classification model initialization with SSL based approach is more effective than the ImageNet weights. In the SSL approach, for the NHS dataset, best accuracy, recall, and F1$\_$Score are obtained when $N = 16$; and best precision is received when $N=8$. In the SSL approach for ALTS dataset, best accuracy and precision are obtained when $N = 16$; best recall and F1$\_$Score are obtained when $N = 8$. According to our experimental results, we observe a noticeable performance improvement in using SSL. Thus the experimental results justify the importance of SSL-based cervical model development. The CSSL based initialization approach further improves the performance of SSL for both datasets. This observation supports the development of the cervix model uniting images from both datasets. We find that for CSSL, in general, $N = 8$ produces the best performance in both NHS and ALTS. Hence, we evaluate the performance of both Federated Self-Supervised Learning (FSSL) models for $N = 8$. For CSFSSL,  we vary the value of $E$ i.e. local model updating for $1$, $5$, $10$ and find that $E = 1$ provides us the best result which is reported in the ninth row of Table~\ref{Table:Performance}. For PPFSSL, we opt two approaches. The first approach starts SSL training with NHS (called PPFSSL$\_$NHS) images and the other approach starts SSL training with ALTS (called PPFSSL$\_$ALTS) images. We set $E = 1$ for experimental similarity, and the classification performance with this model initialization is listed in the tenth and eleventh rows of Table~\ref{Table:Performance}. We find that both PPFSSL and CSFSSL produce comparative performance and in general, federated SSL produces better results than SSL with images in its own dataset. This justifies the effectiveness of FSSL in addressing data sharing constraints in cervix model development.


\begin{table}
\centering
\caption{Performance evaluation}
\resizebox{\linewidth}{!}{
\begin{tabular}{|l|c|c|c|c|c|c|c|c|} 
\hline
\multirow{2}{*}{Initialization } & \multicolumn{4}{c|}{NHS}             & \multicolumn{4}{c|}{ALTS}             \\ 
\cline{2-9}
 Method                                        & ACC     & Recall & Precision & F1\_Score     & ACC    & Recall & Precision & F1\_Score      \\ 
\hline
Random                                   & 79.73 & 0.5306 & 0.7879    & 0.6341 & 76.77 & 0.5735 & 0.6964    & 0.6290  \\ 
\hline
ImageNet                                 & 80.41 & 0.4898 & \textbf{0.8571}    & 0.6234 & 77.78 & 0.5588 & 0.7308    & 0.6333  \\ 
\hline
SSL (N 8)                          & 83.11 & 0.6327 & 0.8158    & 0.7126 & 79.80 & 0.7206 & 0.7000    & 0.7101  \\ 
SSL (N 16)                         & 83.11 & 0.6939 & 0.7727    & 0.7312 & 80.30 & 0.6029 & 0.7736    & 0.6777  \\ 
SSL (N 32)                         & 83.11 & 0.6531 & 0.8000    & 0.7191 & 79.29 & 0.6176 & 0.7368    & 0.6720  \\ 
\hline
CSSL (N 8)                     & \textbf{86.49} & 0.7143 & 0.8537    & 0.7778 & \textbf{81.82} & \textbf{0.7794} & 0.7162    & \textbf{0.7465}  \\ 
CSSL (N 16)                    & 85.81 & 0.7551 & 0.8043    & 0.7789 & 81.31 & 0.6324 & \textbf{0.7818}    & 0.6992  \\ 
CSSL (N 32)                    & 85.14 & \textbf{0.7959} & 0.7647    & \textbf{0.7800} & 81.31 & 0.6471 & 0.7719    & 0.7040  \\
\hline
CSFSSL	&84.46&	0.6735&	0.8250&	0.7416&79.80&	0.6912&	0.7121&	0.7015\\
\hline
PPFSSL$\_$NHS&		84.46&	0.7347&	0.7826&	0.7579&80.81&	0.6618&	0.7500&	0.7031\\
PPFSSL$\_$ALTS	&84.46&	0.7755&	0.7600&	0.7677&80.30&	0.6618&	0.7377&	0.6977\\
\hline

\end{tabular}
}

\label{Table:Performance}
\end{table}

\section{Conclusion and scope of future work}\label{sec: Conclusion}
This paper discusses the challenges behind cervical image analysis due to labeling unavailability and variability and presents a novel direction to address them. Experimental results shows that the self-supervised learning algorithm is proved to be an efficient and effective candidate to deal with the label scarcity as well as the labeling variability.  In addition, the experimentations on Federated Self-Supervised Learning shed into light to deal with the data-sharing restrictions. To the best of our knowledge, this is the first attempt to develop a cervix model in light of a domain-specific pre-trained model for task-specific fine-tuning.

The development of an improved cervix model by uniting larger datasets from different sources independent of labeling criteria, unavailability of labels, and different imaging devices is the immediate future scope of this work. The cervical image datasets for which the labeling efforts made by researchers at NCI are in progress will be used for this research. The engineering implementation of federated learning to work in a real scenario is another important future work. The presented idea can be employed by other medical image analysis tasks for utilizing unlabeled data and providing the same domain transfer learning.

\section{Acknowledgement}\label{sec:acknowledge}
We are very much grateful to Dr. Mark Schiffman of the Division of Cancer Epidemiology and Genetics, National Cancer Institute, National Institutes of Health, and his team for providing us the images and labels used in this paper.

\bibliographystyle{IEEEtran}
\bibliography{refs}

\begin{thebibliography}{10}
\providecommand{\url}[1]{#1}
\csname url@samestyle\endcsname
\providecommand{\newblock}{\relax}
\providecommand{\bibinfo}[2]{#2}
\providecommand{\BIBentrySTDinterwordspacing}{\spaceskip=0pt\relax}
\providecommand{\BIBentryALTinterwordstretchfactor}{4}
\providecommand{\BIBentryALTinterwordspacing}{\spaceskip=\fontdimen2\font plus
\BIBentryALTinterwordstretchfactor\fontdimen3\font minus
  \fontdimen4\font\relax}
\providecommand{\BIBforeignlanguage}[2]{{%
\expandafter\ifx\csname l@#1\endcsname\relax
\typeout{** WARNING: IEEEtran.bst: No hyphenation pattern has been}%
\typeout{** loaded for the language `#1'. Using the pattern for}%
\typeout{** the default language instead.}%
\else
\language=\csname l@#1\endcsname
\fi
#2}}
\providecommand{\BIBdecl}{\relax}
\BIBdecl

\bibitem{Liming2019}
L.~Hu, D.~Bell, S.~Antani, Z.~Xue, K.~Yu, M.~P. Horning, N.~Gachuhi, B.~Wilson,
  M.~S. Jaiswal, B.~Befano, L.~R. Long, R.~Herrero, M.~H. Einstein, R.~D. Burk,
  M.~Demarco, J.~C. Gage, A.~C. Rodriguez, N.~Wentzensen, and M.~Schiffman,
  ``{An Observational Study of Deep Learning and Automated Evaluation of
  Cervical Images for Cancer Screening},'' \emph{JNCI: Journal of the National
  Cancer Institute}, vol. 111, no.~9, pp. 923--932, 01 2019.

\bibitem{Srinivasan2006}
Y.~{Srinivasan}, B.~{Nutter}, S.~{Mitra}, B.~{Phillips}, and E.~{Sinzinger},
  ``Classification of cervix lesions using filter bank-based texture mode,'' in
  \emph{19th IEEE Symposium on Computer-Based Medical Systems (CBMS'06)}, 2006,
  pp. 832--840.

\bibitem{Kim2013}
E.~Kim and X.~Huang, \emph{A Data Driven Approach to Cervigram Image Analysis
  and Classification}.\hskip 1em plus 0.5em minus 0.4em\relax Dordrecht:
  Springer Netherlands, 2013, pp. 1--13.

\bibitem{Fernandes2018}
K.~{Fernandes}, J.~S. {Cardoso}, and J.~{Fernandes}, ``Automated methods for
  the decision support of cervical cancer screening using digital
  colposcopies,'' \emph{IEEE Access}, vol.~6, pp. 33\,910--33\,927, 2018.

\bibitem{Litjens2017}
\BIBentryALTinterwordspacing
G.~Litjens, T.~Kooi, B.~E. Bejnordi, A.~A.~A. Setio, F.~Ciompi, M.~Ghafoorian,
  J.~A. {van der Laak}, B.~{van Ginneken}, and C.~I. Sánchez, ``A survey on
  deep learning in medical image analysis,'' \emph{Medical Image Analysis},
  vol.~42, pp. 60 -- 88, 2017. [Online]. Available:
  \url{http://www.sciencedirect.com/science/article/pii/S1361841517301135}
\BIBentrySTDinterwordspacing

\bibitem{Ching2018}
T.~Ching, D.~S. Himmelstein, B.~K. Beaulieu-Jones, A.~A. Kalinin, B.~T. Do,
  G.~P. Way, E.~Ferrero, P.-M. Agapow, M.~Zietz, M.~M. Hoffman, W.~Xie, G.~L.
  Rosen, B.~J. Lengerich, J.~Israeli, J.~Lanchantin, S.~Woloszynek, A.~E.
  Carpenter, A.~Shrikumar, J.~Xu, E.~M. Cofer, C.~A. Lavender, S.~C. Turaga,
  A.~M. Alexandari, Z.~Lu, D.~J. Harris, D.~DeCaprio, Y.~Qi, A.~Kundaje,
  Y.~Peng, L.~K. Wiley, M.~H.~S. Segler, S.~M. Boca, S.~J. Swamidass, A.~Huang,
  A.~Gitter, and C.~S. Greene, ``Opportunities and obstacles for deep learning
  in biology and medicine.'' \emph{Journal of the Royal Society, Interface},
  April, 2018.

\bibitem{yang2020}
Q.~Yang, Y.~Zhang, W.~Dai, and S.~J. Pan, \emph{Transfer Learning}.\hskip 1em
  plus 0.5em minus 0.4em\relax Cambridge University Press, 2020.

\bibitem{Yang2019}
Q.~{Yang}, Y.~{Liu}, Y.~{Cheng}, Y.~{Kang}, T.~{Chen}, and H.~{Yu}, 2019.

\bibitem{Yang2020Fed}
M.~Yang, A.~Wong, H.~Zhu, H.~Wang, and H.~Qian, ``Federated learning with class
  imbalance reduction,'' 2020.

\bibitem{Zhou2021}
Z.~Zhou, V.~Sodha, J.~Pang, M.~B. Gotway, and J.~Liang, ``Models genesis,''
  \emph{Medical Image Analysis}, vol.~67, p. 101840, 2021.

\bibitem{Chen2020}
T.~Chen, S.~Kornblith, M.~Norouzi, and G.~Hinton, ``A simple framework for
  contrastive learning of visual representations,'' in \emph{Proceedings of the
  37th International Conference on Machine Learning}, ser. Proceedings of
  Machine Learning Research, H.~D. III and A.~Singh, Eds., vol. 119.\hskip 1em
  plus 0.5em minus 0.4em\relax Virtual: PMLR, 13--18 Jul 2020, pp. 1597--1607.

\bibitem{Bratti2004}
M.~C. Bratti, A.~C. Rodríguez, M.~Schiffman, A.~Hildesheim, J.~Morales,
  M.~Alfaro, D.~Guillén, M.~Hutchinson, M.~E. Sherman, C.~Eklund,
  J.~Schussler, J.~Buckland, L.~A~Morera, F.~Cárdenas, M.~Barrantes,
  E.~Pérez, T.~J~Cox, R.~D~Burk, and R.~Herrero, ``Description of a seven-year
  prospective study of human papillomavirus infection and cervical neoplasia
  among 10000 women in guanacaste, costa rica,'' \emph{Pan American journal of
  public health}, vol.~2, no.~15, pp. 75--89, Feb 2004.

\bibitem{Schiffman2000}
M.~Schiffman and M.~E. Adrianza, ``Ascus-lsil triage study. design, methods and
  characteristics of trial participants,'' \emph{Acta Cytol}, vol. 44(5), pp.
  726--742, Sept-Oct 2000.

\bibitem{Chollet2015}
F.~Chollet \emph{et~al.}, ``Keras,'' \url{https://keras.io}, 2015.

\end{thebibliography}
\balance

\end{document}